\documentclass[letterpaper, 10 pt, conference]{ieeeconf}  

\IEEEoverridecommandlockouts                              

\usepackage{amsmath,amsfonts}
\usepackage{algorithmic}
\usepackage{algorithm}
\usepackage{array}
\usepackage[caption=false,font=normalsize,labelfont=sf,textfont=sf]{subfig}
\usepackage{textcomp}
\usepackage{stfloats}
\usepackage{url}
\usepackage{verbatim}
\usepackage{graphicx}
\usepackage{cite}
\usepackage{multirow}
\usepackage{makecell}
\usepackage[hidelinks]{hyperref}
\usepackage{xcolor}
\usepackage{balance}
\usepackage{xurl}
\usepackage[hidelinks]{hyperref}

\newcommand{\tblfont}{\fontsize{7.8pt}{9.0pt}\selectfont}
\newcommand{\tblcolsep}{4.0pt}
\newcommand{\tblstretch}{1.5}

\title{\LARGE \bf
LCGNav: Local Candidate-Aware Geometric Enhancement for General Topological Planning in Vision-Language Navigation
}

\author{Jiankun Peng$^{1,2}$, Jianyuan Guo$^{3}$, Yiguang Yang$^{1,2}$, Yue Liu$^{1,2}$, Jiashuang Yan$^{1,2}$, Ying Xu$^{1}$
\thanks{$^{1}$ The Aerospace Information Research Institute, Chinese Academy of Sciences, Beijing 100094, China
        {\tt\small }}%
\thanks{$^{2}$ The School of Electronic, Electrical and Communication Engineering, University of Chinese Academy of Sciences, Beijing 100049, China
        {\tt\small }}%
\thanks{$^{3}$ The Department of Computer Science, City University of Hong Kong, Hong Kong, SAR, China
        {\tt\small }}%
}

\begin{document}

\maketitle
\thispagestyle{empty}
\pagestyle{empty}

\begin{abstract}
Online topological planning has become an effective paradigm for Vision-Language Navigation in Continuous Environments (VLN-CE), but existing methods still suffer from two limitations: redundant local depth information and weakened focus on current frontier candidates as the topological graph grows. To address this, we propose LCGNav, a modular local geometric enhancement framework for topological VLN. LCGNav explicitly converts candidate depth views into 3D point clouds and applies physical truncation based on the agent’s reachable range, enabling more compact local geometric modeling. It further introduces a dimension-preserving local fusion strategy with transient state degradation, so that geometric enhancement is applied only to the currently relevant ghost nodes without changing the original planner interface. Experiments on R2R-CE and RxR-CE show that LCGNav serves as an effective cross-architecture enhancement module, consistently improving multiple key metrics of representative online topological baselines with low additional training cost. When integrated with ETP-R1, LCGNav achieves the best performance among the compared online topological methods on the val-unseen splits of the R2R-CE and RxR-CE benchmarks. The code is available at 
\href{https://github.com/shannanshouyin/LCGNav}
{\nolinkurl{https://github.com/shannanshouyin/LCGNav}}. 
\end{abstract}


\section{INTRODUCTION}

Vision-Language Navigation (VLN) requires an embodied agent to navigate to a target location by following natural language instructions\cite{anderson2018vision}. While early VLN studies mainly focused on discrete environments with predefined navigation graphs, recent work has increasingly shifted to Vision-Language Navigation in Continuous Environments (VLN-CE), where the agent must handle low-level continuous control, obstacle avoidance, and long-horizon spatial reasoning without oracle connectivity\cite{krantz2020beyond}.

Existing VLN-CE methods mainly follow two directions: end-to-end learning and map-based hierarchical planning. End-to-end methods directly predict actions from observations, but often struggle with catastrophic forgetting and spatial disorientation over long trajectories due to the lack of structured memory\cite{chen2021history}, \cite{zhang2024navid}, \cite{zhang2024uni}, \cite{wei2025streamvln}. By contrast, map-based hierarchical planning introduces explicit spatial memory and decouples high-level planning from low-level control\cite{chen2021topological}, \cite{chen2022think}, \cite{georgakis2022cross}. Within this line, online topological planning has proven particularly effective. For example, ETPNav~\cite{an2024etpnav} constructs a topological graph on the fly, where unvisited candidate waypoints at the local frontier are represented as \emph{ghost nodes}. This design provides persistent spatial memory for exploration in unseen environments.

Despite their success, we argue that there remains room to improve how local frontier candidates are represented and used during online topological planning. In particular, effective candidate modeling should satisfy two requirements. First, it should emphasize the geometry that is physically relevant to the agent's immediate reachable space. Raw depth observations often include distant regions that are weakly related to the next waypoint decision, while directly truncating them in the 2D image domain may break the continuity of depth observations and introduce unnatural local patterns for the visual backbone. We therefore perform truncation after projecting candidate depth views into 3D space, where distant geometry can be removed in a physically meaningful manner while better preserving local structure. Second, candidate enhancement should remain local and transient rather than being accumulated throughout the global topological graph. As the graph grows, historical nodes may dilute the planner's attention to the current frontier, especially at complex intersections. We thus restrict geometric enhancement to currently relevant ghost nodes and degrade the enhanced states after they become historical, enabling the planner to focus on the current action space.

Motivated by these two design principles, we propose LCGNav, a modular Local Candidate-Aware Geometric Enhancement framework for VLN-CE. LCGNav explicitly converts candidate depth views into 3D point clouds and applies physical truncation based on the agent's reachable range, enabling compact local geometric modeling. It further introduces a dimension-preserving local fusion strategy with transient state degradation, so that geometric enhancement is applied only to currently relevant ghost nodes without changing the original planner interface.

In summary, our contributions are three-fold. First, we propose LCGNav, a modular geometric enhancement framework that can be integrated into existing online topological navigation methods with strong cross-architecture transferability. Second, we introduce an explicit 3D truncation strategy and a dimension-preserving local fusion mechanism with transient state degradation to reduce redundant depth information and mitigate the influence of historical nodes on current local decision-making. Third, experiments on the R2R-CE and RxR-CE benchmarks show that LCGNav improves multiple key metrics across representative baselines, and achieves the best performance among the compared online topological methods when integrated with ETP-R1 on the corresponding val-unseen splits.

\section{Related Work}

\subsection{Vision-Language Navigation in Continuous Environments}

Unlike discrete VLN\cite{anderson2018vision}, Vision-Language Navigation in Continuous Environments (VLN-CE) requires agents to follow natural language instructions under low-level continuous control\cite{krantz2020beyond}, making obstacle avoidance and long-horizon reasoning substantially more challenging. Existing VLN-CE methods mainly fall into two categories: end-to-end learning and map-based hierarchical planning. End-to-end methods directly predict control signals from multimodal observations and have recently benefited from LLM/LVLM-based reasoning, but they still rely heavily on implicit memory and often struggle in long trajectories\cite{chen2021history}, \cite{zhang2024navid}, \cite{zhang2024uni}, \cite{wei2025streamvln}. By contrast, map-based hierarchical methods build explicit spatial memory, such as metric or topological maps\cite{georgakis2022cross}, \cite{chen2021topological}, \cite{wang2023gridmm}, and decouple high-level planning from low-level control, which better supports long-horizon navigation. Our work follows this map-based line and focuses on improving geometric perception and feature integration at local spatial frontiers.

\subsection{Online Topological Planning}

Online topological planning provides a structured alternative to directly predicting low-level actions in continuous VLN. These methods predict locally reachable waypoints from the current observation and organize both visited locations and unvisited waypoint candidates into a dynamically updated topological graph~\cite{chen2022think,hong2022bridging}. The unvisited candidates, often referred to as \emph{ghost nodes}, define the current high-level action space, from which a cross-modal planner selects the next target according to the language instruction and accumulated graph memory. Following this paradigm, ETPNav~\cite{an2024etpnav} constructs an evolving topological graph for VLN-CE, and subsequent methods further improve it through dynamic topology modeling (DGNav~\cite{peng2026dynamic}), collision-aware waypoint prediction (Safe-VLN~\cite{yue2024safe}), improved cross-modal grounding (WP-CMA~\cite{fu2025wp}), and stronger training with LLM-generated trajectories and reinforcement fine-tuning (ETP-R1~\cite{ye2025etp}). Despite these advances, most existing topological methods still rely on 2D depth observations to represent frontier candidates. Because the agent's immediate motion range is bounded, raw depth maps often contain substantial distant redundancy, and direct truncation in the image domain may break local depth continuity and introduce unnatural patterns for the visual backbone. These limitations motivate our explicit 3D geometric modeling of local frontier candidates.

\subsection{Spatial Perception and Geometric Representations}

Spatial representation in VLN has been studied through both explicit maps and implicit memory. Explicit approaches, such as GridMM\cite{wang2023gridmm}, CM2\cite{georgakis2022cross}, BEVBert\cite{an2022bevbert}, and OVL-MAP\cite{wen2025ovl}, construct top-down or hybrid spatial maps to model environment structure, while implicit approaches such as JanusVLN reduce mapping overhead by encoding history in latent memory\cite{zeng2025janusvln}. Although dense metric maps provide rich spatial semantics, they often introduce higher computation and weaker transferability, whereas implicit methods may lack stable long-horizon structure. Online topological planning offers a practical middle ground, especially when panoramic RGB-D observations are available. However, existing topological methods still underuse this geometric information at local frontiers by directly encoding raw 2D depth views. In contrast, our method explicitly converts local depth observations into truncated 3D point clouds to provide more compact local geometric cues without introducing dense metric mapping overhead.

\begin{figure*}
    \centering
    \includegraphics[width=0.95\linewidth]{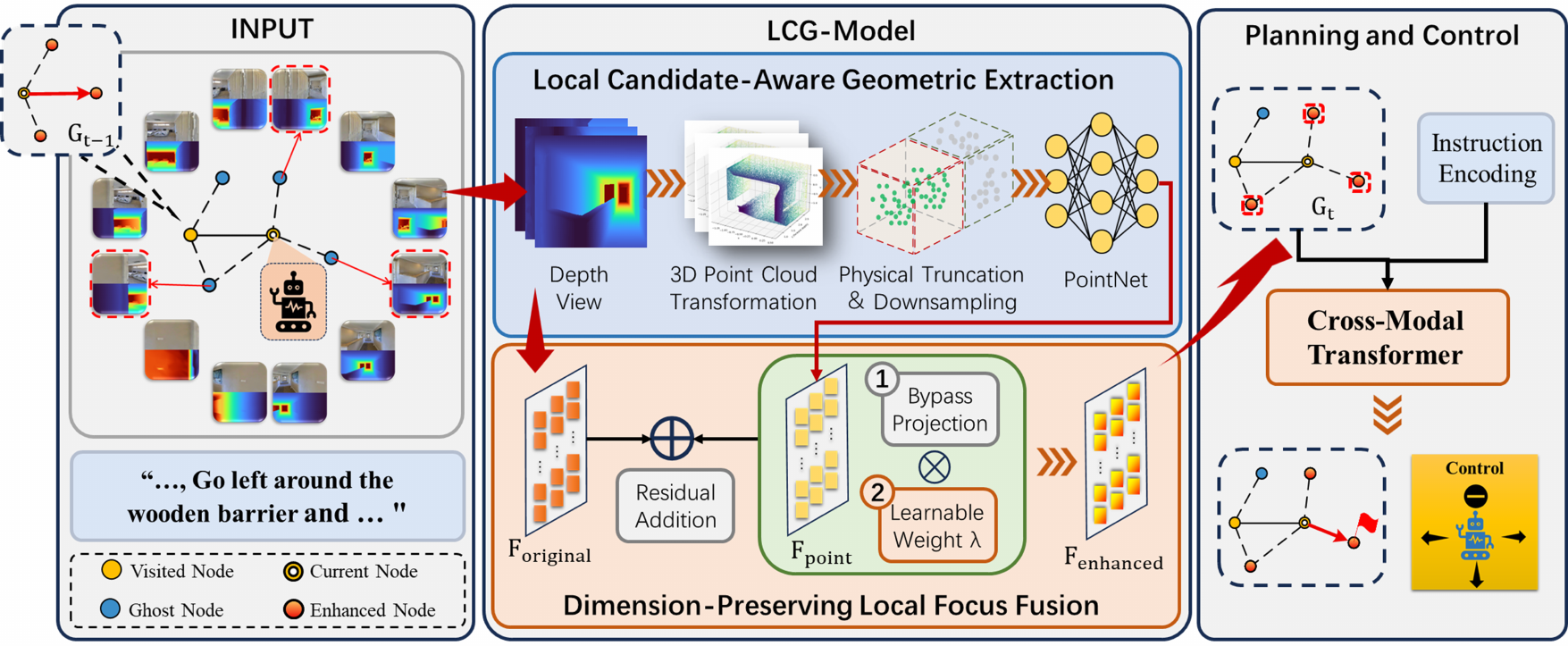}
    \caption{The overall framework of LCGNav. At each step, the agent generates local frontier candidates (ghost nodes) from the 12 panoramic observations. The depth views corresponding to these candidates are explicitly converted into 3D point clouds and physically truncated based on the maximum reachable distance to extract local geometric features. Subsequently, these enhanced features are integrated into the dynamically evolving topological graph, guiding the cross-modal planner to focus on the current action space for navigation.}
\end{figure*}

\section{Method}

\subsection{Problem Formulation and LCGNav Overview}

\textbf{Task Setup}. In Vision-Language Navigation in Continuous Environments (VLN-CE), an embodied agent is required to navigate through an unknown continuous 3D environment by following a natural language instruction, which is tokenized as $I=\{w_1,w_2,\cdots,w_L\}$, where $L$ denotes the number of tokens.
At each time step t, the agent perceives an egocentric panoramic observation $O_t=\left\{V_t,D_t,p_t\right\}$, which consists of 12 discrete views spanning 360 degrees. Here, $V_t$ and $D_t$ denote the sets of RGB images and depth maps across these 12 views, respectively, while $p_t=\left(x_t,y_t,\theta_t\right)$ represents the current agent pose provided by an internal pose sensor. Unlike discrete VLN where the agent teleports between predefined graph nodes, VLN-CE requires the agent to move within a continuous physical space. The low-level action space is defined as $\mathcal{A}=  \{ \text{MOVEFORWARD}(0.25\,\text{m}), \text{TURNLEFT}/\text{TURNRIGHT} \\ (15^\circ), \text{STOP} \}$. The navigation is considered successful if the Euclidean distance between the agent's final position and the target location is within a predefined threshold $d_{th}=3.0m$ when the STOP action is executed.

\textbf{Online Topological Mapping Paradigm}. To support long-horizon reasoning in continuous environments, we build on the Evolving Topological Planning (ETP) paradigm~\cite{an2024etpnav,peng2026dynamic,ye2025etp}. At each step, the agent maintains a dynamically updated topological graph $\mathcal{G}_t=(\mathcal{V}_t,\mathcal{E}_t)$, where each node stores the panoramic observation of a physical location. The node set consists of three types: the current node $\mathcal{V}_{\mathrm{cur}}$, visited nodes $\mathcal{V}_{\mathrm{vis}}$, and unvisited ghost nodes $\mathcal{V}_{\mathrm{ghost}}$, which represent candidate waypoints at the local frontier. The planning cycle proceeds as follows. First, a waypoint predictor processes the current observation $O_t$ and proposes navigable local waypoints, which are incorporated into the graph as ghost nodes connected to the current node. Previously generated ghost nodes remain in the graph until they are visited or invalidated by subsequent graph updates. Second, node representations from the graph, together with the instruction $I$, are fed into a transformer-based cross-modal planner, which predicts the most promising ghost node as the next target $v_{\mathrm{target}}$. Finally, a low-level obstacle-avoiding controller executes continuous actions to move toward $v_{\mathrm{target}}$. Once the target is reached, the selected ghost node becomes a visited node, and the process repeats.

Existing ETP-style methods typically construct ghost-node spatial features directly from raw 2D depth observations. Because the agent's immediate motion range is bounded, such observations often contain substantial distant redundancy, while direct truncation in the image domain may disrupt local depth continuity and introduce unnatural patterns for the visual backbone. Our method addresses this limitation by enhancing ghost-node representations with explicit local 3D geometry.

\textbf{LCGNav Overview}. To address the limitations of raw 2D depth representations and the resulting global attention dilution, we propose the Local Candidate-Aware Geometric Enhancement (LCGNav) framework. As illustrated in Figure 1, LCGNav functions as a modular enhancement integrated before the cross-modal planner. Rather than directly feeding the raw 2D depth views of candidate nodes into the visual backbone, LCGNav explicitly projects these views into 3D point clouds. By applying spatial truncation based on the agent’s physical reachable distance, it filters out distant spatial redundancy and extracts local geometric features using a PointNet encoder. Subsequently, a Dimension-Preserving Local Focus Fusion mechanism is introduced to inject these geometric features into the corresponding ghost node tokens within $\mathcal{G}_t$ via bypass projection and residual addition. Furthermore, this enhancement is transient: once the agent visits a ghost node, its enriched representation is reset to the standard topological feature. This design acts as a local spotlight, helping the planner focus on the current action space while preserving the original topological feature dimensions and graph structure.

\subsection{Local 3D Geometric Perception via Physical Truncation}

\textbf{Explicit 3D Point Cloud Transformation}. As analyzed previously, directly extracting spatial features from 2D depth maps may introduce substantial redundant information from distant, unreachable regions. To improve local spatial perception, LCGNav isolates the depth view corresponding to each newly generated ghost node $\mathbf{v}_{\text{ghost}}^j \in \mathbf{V}_{\text{ghost}}$ and explicitly transforms it into a 3D point cloud. Given a local depth map $D_j\in\mathbf{R}^{\mathbf{H}\times\mathbf{W}}$ associated with the j-th candidate view, any valid pixel coordinate $\left(u,\ v\right)$ with a depth value $z=D_j\left(u,\ v\right)$ is projected into the 3D camera coordinate system using the camera intrinsic matrix K. The corresponding 3D spatial coordinate $p_i=\left(x_i,y_i,z_i\right)$ is computed as follows:

\begin{equation}
\label{deqn_ex1a}
x_i=\frac{\left(u-c_x\right)\bullet z}{f_x},\ y_i=\frac{\left(v-c_y\right)\bullet z}{f_y},\ z_i=z.
\end{equation}

where $f_x$, $f_y$ denote the focal lengths and $\left(c_x,c_y\right)$ is the principal point. By applying this projection to all valid pixels in the depth map, we obtain the raw dense 3D point cloud ${PC}_{raw}^j=\left\{p_1,p_2,\cdots,p_K\right\}$ for the candidate view.

\textbf{Depth-directed Physical Hard Truncation}. Indiscriminately applying pixel-level hard masking on 2D depth maps disrupts the local smoothness of the image and may introduce sharp truncation artifacts in the feature space, which is detrimental to CNN or ViT backbones that rely on continuous receptive fields. In contrast, operating within an explicit 3D coordinate system allows us to implement spatial cropping with clear physical significance. Considering that the agent's immediate physical action range is strictly bounded, we define a maximum reachable depth threshold $d_{max}=3.0m$ in this work, aligning with the local waypoint generation range. Crucially, rather than employing a panoramic Euclidean distance-based spherical cutoff, we perform hard truncation strictly along the depth direction (i.e., the Z-axis) of the camera view. By applying this depth-based spatial filter, we better isolate the effective physical interaction space directly in front of the agent while stripping away distant background noise:

\begin{equation}
\label{deqn_ex1a}
{PC}_{trunc}^j=\left\{p_i\in{PC}_{raw}^j\ |\ \ z_i\le d_{max}\right\}.
\end{equation}

This physical truncation operation helps reduce geometric redundancy at the far end of the view frustum without compromising the integrity of the local 3D structure. The effectiveness of this 3D physical truncation, compared with direct 2D cropping, is validated in Table~\ref{tab:ablation_repr_scope}.

\textbf{Farthest Point Downsampling (FPS)}. The truncated point cloud ${PC}_{trunc}^j$ contains a dynamic number of points and remains too dense for lightweight processing. To balance computational efficiency and provide structured input of fixed dimensions for the subsequent neural network, we introduce the Farthest Point Sampling (FPS) algorithm. FPS helps preserve the overall geometric topology and boundary features while downsampling ${PC}_{trunc}^j$ into a uniformly distributed, fixed-size point set:

\begin{equation}
\label{deqn_ex1a}
{PC}_{final}^j=FPS\left({PC}_{trunc}^j,\ N_{pts}\right).
\end{equation}

where $N_{pts}$ is the predefined number of points ($N_{pts}=256$ in our experiments). The resulting point cloud ${PC}_{final}^j\in\mathbf{R}^{N_{pts}\times3}$, free from distant interference, provides a compact geometric representation of the candidate node's local action space, which is then fed into the network for geometric feature extraction.

\begin{figure}[!t]
\centering
\includegraphics[width=3.5in]{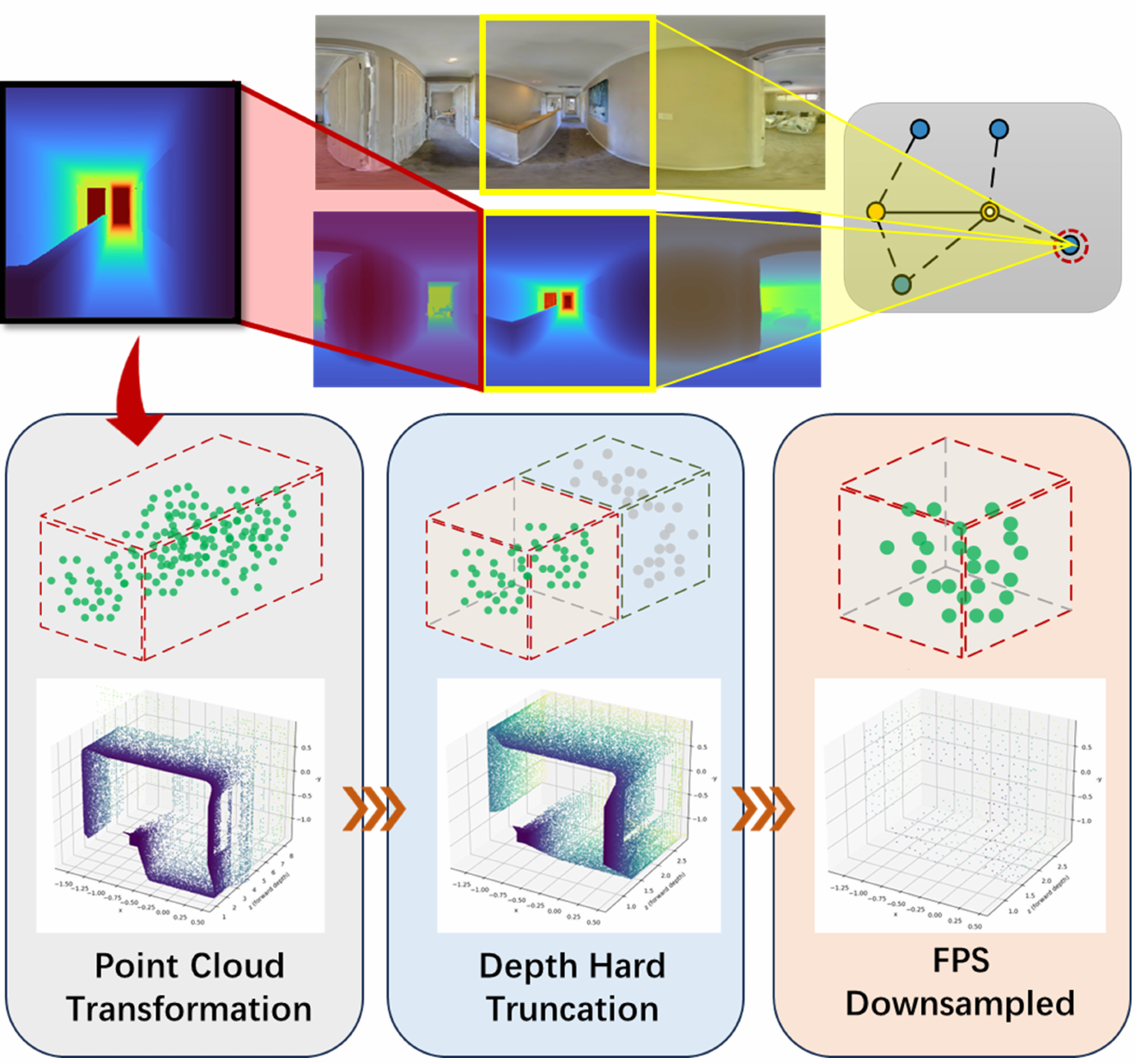}
\caption{Local 3D Geometric Perception via Physical Truncation. The local depth view of a candidate ghost node is explicitly projected into a 3D point cloud using camera intrinsics. A physical hard truncation ($d_{max}=3.0m$) is applied strictly along the Z-axis to reduce distant spatial redundancy while avoiding direct 2D truncation artifacts. Finally, Farthest Point Sampling (FPS) downsamples the truncated point cloud to generate a fixed-dimensional geometric representation.}
\label{fig_2}
\end{figure}

\subsection{Dimension-Preserving Local Focus Fusion}

\textbf{Point Cloud Feature Extraction}. The physically truncated point $cloud {PC}_{final}^j$ is fed into a PointNet encoder to extract local geometric features. Specifically, the PointNet processes the 3D coordinates through a shared Multi-Layer Perceptron (MLP) to extract high-dimensional point-wise representations. To ensure stable training and feature scaling, Batch Normalization is applied with no dropout. A max-pooling operation is subsequently employed to aggregate the pointwise features, yielding a dense, global point cloud representation $g_j\in\mathbb{R}^D$, where D matches the hidden dimension of the visual backbone. We emphasize that the goal of this design is not to replace the 2D visual backbone with a stronger standalone 3D encoder. Instead, the point-cloud branch serves as a lightweight geometric complement that injects physically grounded local 3D cues into ghost-node representations after removing distant redundancy.

\textbf{Dimension-Preserving Residual Fusion}. Rather than indiscriminately modifying the global topological map, the extracted geometric feature $g_j$ is exclusively fused with its corresponding local candidate node. To align the feature spaces, $g_j$ is processed through a bypass projection matrix $\mathbf{W}_{proj}$ and scaled by a learnable scalar weight $\lambda$. The resulting weighted point cloud feature is then integrated directly with the original ghost image feature $v_{ghost}^j\in\mathbf{R}^D$ via a residual addition:

\begin{equation}
\label{deqn_ex1a}
v_{enhanced}^j=v_{ghost}^j+\lambda\bullet\left(\mathbf{W}_{proj}\bullet g_j\right).
\end{equation}

This residual fusion preserves the original visual token dimension in the enhanced ghost feature. By avoiding structural disruptions to the token sequence, it maintains architectural compatibility while introducing additional local geometric cues to the cross-modal planner. We later verify the necessity of this dimension-preserving weighted fusion through dedicated comparisons with direct fusion in Table~\ref{tab:ablation_fusion}.

\textbf{State Degradation Mechanism.} To prevent the accumulation of geometric features from biasing the global attention distribution over long horizons, we introduce a transient state degradation strategy. Crucially, the geometric enhancement is applied only to the ghost nodes connected to the agent's current node ($v_{curr}$), acting as a “local spotlight” for the immediate action space. When the agent executes an action and transitions to a target node, the graph updates and a dual degradation occurs: (1) the selected target node transitions from a ghost node to the new current node, and (2) the remaining unselected ghost nodes associated with the previous location become obsolete frontiers. For both cases, their injected 3D geometric vectors are systematically discarded, automatically degrading their enriched representations back to standard topological features. This strategy reduces the influence of historical trajectories and helps the planner focus on the current exploratory space. The benefit of restricting enhancement to current ghost nodes, together with transient state degradation, is further validated by comparing local and global enhancement settings in in Table~\ref{tab:ablation_repr_scope}.

\begin{figure}[!t]
\centering
\includegraphics[width=3.5in]{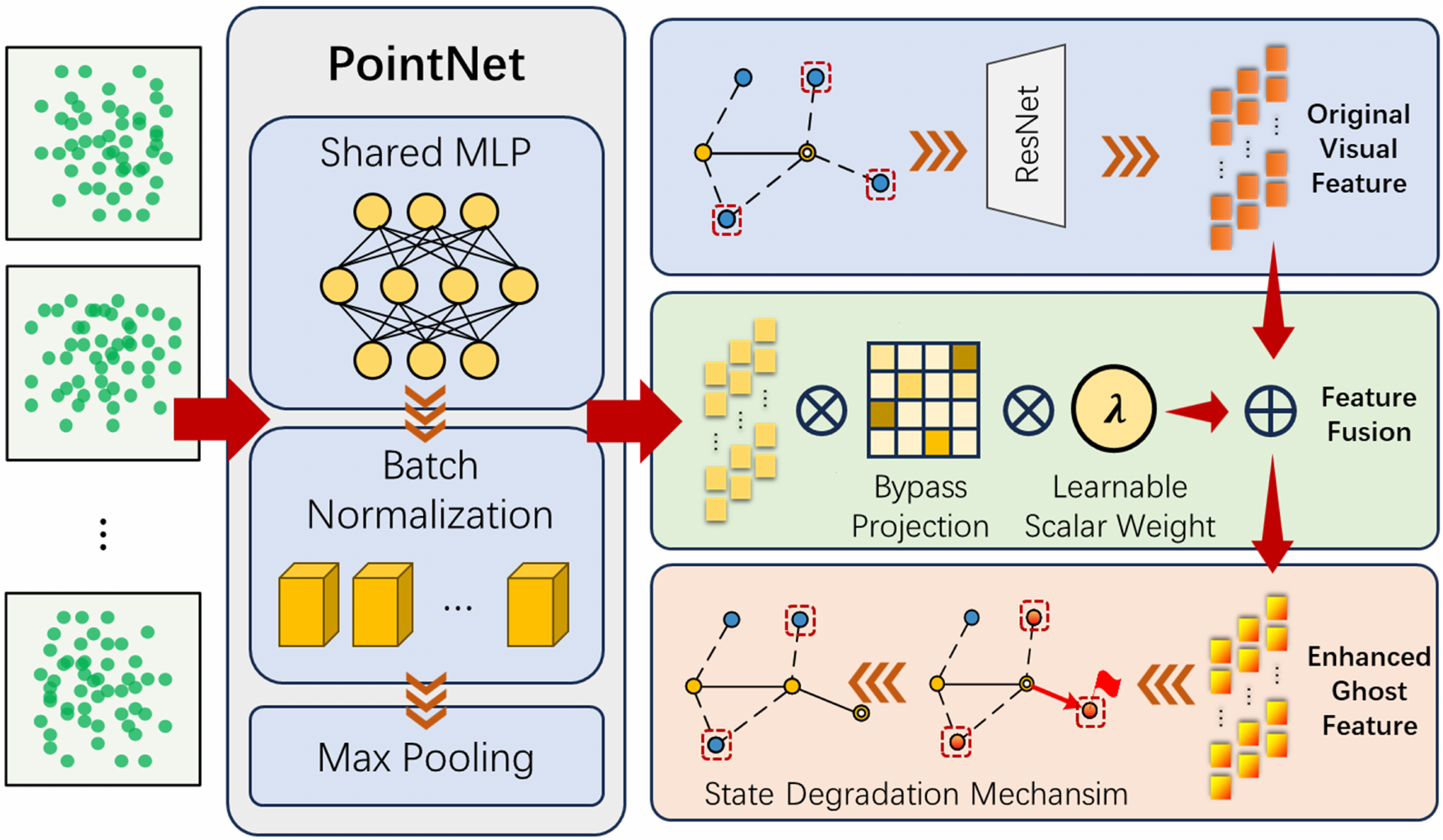}
\caption{ Dimension-Preserving Local Focus Fusion with State Degradation Mechanism. Local geometric features extracted via a PointNet encoder are residually fused with their corresponding ghost image tokens while preserving the original token dimensions. A transient state degradation strategy subsequently discards these injected 3D vectors once the nodes are visited or bypassed, reducing the influence of historical trajectories and helping the planner focus on the current exploratory frontiers.}
\label{fig_3}
\end{figure}

\section{Experiments}

\subsection{Experimental Setup}
\textbf{Datasets}. We evaluate LCGNav on two VLN-CE benchmarks, R2R-CE and RxR-CE, both built on Matterport3D (MP3D) and simulated in Habitat\cite{savva2019habitat}. R2R-CE features concise English instructions and shorter shortest-path trajectories, whereas RxR-CE is more challenging, with substantially longer trajectories, multilingual instructions, and stronger path-following requirements.. Following the standard protocol, we report results on the val-unseen split of both benchmarks.

\textbf{Evaluation Metrics}. We adopt the standard VLN-CE metrics, including Trajectory Length (TL), Navigation Error (NE), Oracle Success Rate (OSR), Success Rate (SR), Success weighted by Path Length (SPL), normalized Dynamic Time Warping (nDTW), and Success weighted by nDTW (SDTW). TL measures path length, NE measures the final distance to the goal, OSR indicates whether the trajectory ever reaches the goal region, SR measures whether the agent successfully stops within the target radius, SPL balances navigation success and path efficiency, and nDTW/SDTW evaluate the fidelity between predicted and reference trajectories. Following common practice, we primarily focus on SR and SPL on R2R-CE, while on RxR-CE we emphasize nDTW and SDTW, where adherence to the annotated path is more critical.

\textbf{Implementation Details}. Our LCGNav framework is implemented using PyTorch and trained on two NVIDIA RTX 4090 GPUs for approximately 3 hours. Designed as a modular enhancement, LCGNav is integrated directly before the cross-modal planner without altering the original input dimensions or the underlying architecture of the selected pre-trained navigation baseline. Regarding the specific geometric configurations, the local point cloud for each ghost node is truncated at a depth threshold of 3m, aligning with the agent’s maximum single-step movement distance, and subsequently downsampled to 256 points. The dedicated geometric extractor processes these 3-dimensional input coordinates through a Multi-Layer Perceptron (MLP) with consecutive channel dimensions of [64, 128, 256], utilizing Batch Normalization to stabilize the feature scaling. During the training phase, we adopt a parameter-efficient tuning strategy: all parameters of the baseline model are frozen, and only the newly introduced components—specifically the PointNet, the bypass projection matrix, and the learnable scalar weight—are optimized. The additional training stage is parameter-efficient and requires no more than 1000 iterations (varying slightly depending on the convergence behavior of the specific baseline). To ensure fair comparison with the chosen baseline model, the batch size and optimizer configurations follow its default settings. This brief training phase enables the model to acquire additional 3D geometric perception capabilities while preserving the established navigation and language-grounding logic of the baseline.

\subsection{ Comparison with State-of-the-Art Methods}

\begin{table}[t]
\caption{Comparison with SOTA methods on the R2R-CE dataset.}
\label{tab:r2rce_sota}
\centering
{\tblfont
\setlength{\tabcolsep}{\tblcolsep}
\renewcommand{\arraystretch}{\tblstretch}
\begin{tabular}{@{}c l c c c c c@{}}
\hline
\multicolumn{1}{c}{} & \multirow{2}{*}{Methods} & \multicolumn{5}{c}{Val-Unseen} \\
\cline{3-7}
 &  & TL & NE$\downarrow$ & OSR$\uparrow$ & SR$\uparrow$ & SPL$\uparrow$ \\
\hline
\multirow{9}{*}{\makecell{End\\to\\End}}
 & Seq2Seq\cite{krantz2020beyond}      &  8.64 & 7.37 & 40 & 32 & 30 \\
 & Sim2Sim\cite{krantz2022sim}         & 10.69 & 6.07 & 52 & 43 & 36 \\
 & CWP-CMA\cite{hong2022bridging}      & 10.90 & 6.20 & 52 & 41 & 36 \\
 & CWP-RecBert\cite{hong2022bridging}  & 12.23 & 5.74 & 53 & 44 & 39 \\
 & NaVid\cite{zhang2024navid}          &  7.63 & 5.47 & 49 & 37 & 36 \\
 & Uni-NaVid\cite{zhang2024uni}        &  9.71 & 5.58 & 53 & 47 & 43 \\
 & StreamVLN\cite{wei2025streamvln}    &  --   & 5.43 & 63 & 53 & 47 \\
 & BudVLN\cite{he2026nipping}          &  --   & 4.74 & 66 & 58 & 51 \\
 & JanusVLN\cite{zeng2025janusvln}     &  --   & 4.78 & 65 & 61 & 57 \\
\hline
\multirow{8}{*}{\makecell{Map\\based}}
 & CM2\cite{georgakis2022cross}        & 11.54 & 7.02 & 41.5 & 34.3 & 27.6 \\
 & WG-MGMap\cite{chen2022weakly}       & 10.00 & 6.28 & 48   & 39   & 34   \\
 & MapNav\cite{zhang2025mapnav}        &  --   & 4.93 & 53   & 40   & 37   \\
 & GridMM\cite{wang2023gridmm}         & 13.36 & 5.11 & 61   & 49   & 41   \\
 & DUET-CE\cite{chen2022think}         & 13.08 & 5.16 & 62   & 54   & 46   \\
 & OVL-MAP\cite{wen2025ovl}            & 11.45 & 4.69 & 65   & 60   & 50   \\
 & WP-CMA\cite{fu2025wp}               & 12.95 & 4.52 & 66   & 59   & 50   \\
 & \textbf{LCG-ETP-R1 (ours)}          & 11.84 & \textbf{3.93} & \textbf{71} & \textbf{65} & \textbf{57} \\
\hline
\end{tabular}
}
\end{table}

\textbf{Quantitative Comparison on R2R-CE}. As shown in Table 1, we evaluate the proposed LCGNav, integrated with the ETP-R1\cite{ye2025etp} baseline and denoted as LCG-ETP-R1, against a wide range of representative methods on the R2R-CE Val-Unseen split. The compared methods are categorized into end-to-end approaches and map-based hierarchical methods.

Compared with many traditional end-to-end methods, map-based hierarchical approaches generally show advantages in continuous environments. By constructing explicit spatial memory (e.g., topological or metric maps), they can better support long-horizon spatial reasoning and mitigate catastrophic forgetting during complex exploration. Consistent with this trend, established map-based models such as OVL-MAP\cite{wen2025ovl} and WP-CMA\cite{fu2025wp} achieve SRs of 60\% and 59\%, with SPLs of 50\%, respectively, outperforming several standard end-to-end baselines in Table 1. At the same time, recent end-to-end models with stronger streaming or history mechanisms, such as BudVLN\cite{he2026nipping} and JanusVLN\cite{zeng2025janusvln}, also achieve competitive results. For example, JanusVLN reports an SR of 61\% and an SPL of 57\%, although its NE remains higher than that of our method.

By augmenting the ETP-R1 baseline with our local 3D geometric perception module, LCG-ETP-R1 achieves the strongest overall performance among the compared online topological methods in Table 1. Specifically, our method achieves an SR of 65\% and an OSR of 71\%, indicating strong ability to both explore and locate the target region. Most notably, our approach reduces the NE to 3.93, making it the only compared method in Table 1 with NE below 4.0 and improving over WP-CMA (4.52) and JanusVLN (4.78). Achieving an SPL of 57\% alongside a competitive trajectory length (TL of 11.84) suggests that the proposed geometric enhancement helps improve path efficiency while maintaining strong navigation success.

\begin{table*}[t]
\caption{Cross-Architecture Transferability Evaluation on R2R-CE and RxR-CE.}
\label{tab:cross_arch_transfer_both}
\centering
\setlength{\tabcolsep}{7.0pt}
\renewcommand{\arraystretch}{1.5}
\begin{tabular}{c c c c c c c c c c c c c}
\hline
\multirow{2}{*}{Method} 
& \multirow{2}{*}{Iters} 
& \multicolumn{5}{c}{R2R-CE (Val-Unseen)} 
& \multirow{2}{*}{Iters} 
& \multicolumn{5}{c}{RxR-CE (Val-Unseen)} \\
\cline{3-7} \cline{9-13}
&  & TL & NE$\downarrow$ & OSR$\uparrow$ & SR$\uparrow$ & SPL$\uparrow$
&  & NE$\downarrow$ & SR$\uparrow$ & SPL$\uparrow$ & nDTW$\uparrow$ & SDTW$\uparrow$ \\
\hline

ETPNav\cite{an2024etpnav} & \multirow{2}{*}{60}
& 11.64 & 4.74 & 64.38 & 57.47 & 49.34
& \multirow{2}{*}{350}
& 5.97 & 53.40 & 43.58 & 60.74 & 43.88 \\
LCG-ETP
& 
& 11.68 & \textbf{4.60} & \textbf{65.52} & \textbf{58.07} & \textbf{50.14}
&
& \textbf{5.89} & \textbf{54.00} & \textbf{44.28} & \textbf{61.17} & \textbf{44.43} \\
\hline

BEVBert\cite{an2022bevbert} & \multirow{2}{*}{900}
& 13.28 & 4.62 & 67.16 & 59.22 & 49.21
& \multirow{2}{*}{--}
& -- & -- & -- & -- & -- \\
LCG-BEV
&
& \textbf{11.64} & 4.64 & 66.34 & 58.99 & \textbf{50.92}
&
& -- & -- & -- & -- & -- \\
\hline

DGNav\cite{peng2026dynamic} & \multirow{2}{*}{600}
& 11.60 & 4.72 & 64.65 & 57.74 & 49.54
& \multirow{2}{*}{200}
& 6.00 & 53.78 & 44.37 & 62.04 & 44.49 \\
LCG-DG
&
& 12.43 & 4.70 & \textbf{66.34} & \textbf{59.16} & \textbf{49.61}
&
& \textbf{5.88} & \textbf{54.02} & \textbf{44.88} & \textbf{62.89} & \textbf{45.21} \\
\hline

ETP-R1(SFT)\cite{ye2025etp} & \multirow{2}{*}{500}
& 13.34 & 4.11 & 68.62 & 63.24 & 54.30
& \multirow{2}{*}{150}
& 5.42 & 58.26 & 48.19 & 63.78 & 48.53 \\
LCG-R1(SFT)
&
& \textbf{12.56} & \textbf{4.07} & \textbf{69.28} & \textbf{63.89} & \textbf{55.21}
&
& \textbf{5.38} & \textbf{58.60} & \textbf{48.63} & \textbf{64.13} & \textbf{49.07} \\
\hline

ETP-R1(RFT)\cite{ye2025etp} & \multirow{2}{*}{800}
& 12.74 & 3.94 & 71.45 & 65.31 & 55.77
& \multirow{2}{*}{50}
& 5.25 & 59.39 & 48.60 & 65.06 & 49.97 \\
LCG-R1(RFT)
&
& \textbf{11.84} & \textbf{3.93} & 70.69 & 65.14 & \textbf{56.85}
&
& \textbf{5.22} & \textbf{59.81} & \textbf{48.91} & \textbf{65.38} & \textbf{50.34} \\
\hline

\end{tabular}
\end{table*}

\textbf{Cross-Architecture Transferability}. A key advantage of our method is its modular design, which allows LCGNav to be integrated into different online topological planners with minimal tuning. To evaluate this transferability, we plug the LCG module into representative baselines and optimize only the newly introduced parameters for at most 1000 iterations. As shown in Table 2, LCGNav consistently improves performance across architectures on both R2R-CE and RxR-CE. For baselines without publicly available RxR-CE checkpoints, we report only R2R-CE results.

On R2R-CE, LCGNav brings consistent but architecture-dependent gains. For ETPNav and DGNav, the improvements are mainly reflected in better success-related metrics, suggesting that local 3D cues help distinguish more reliable frontier candidates. For BEVBert and ETP-R1, the gains are more evident in path efficiency, such as shorter trajectories and higher SPL, indicating that the proposed enhancement can reduce redundant exploration even when the baseline already has strong map-based reasoning. More importantly, on the more challenging RxR-CE benchmark, LCGNav improves all reported metrics for every baseline with available checkpoints, including NE, SR, SPL, nDTW, and SDTW. This indicates that the proposed module not only improves goal-reaching performance, but also helps the agent follow instruction-aligned trajectories more faithfully.

These results demonstrate that LCGNav is transferable not only across planner architectures, but also across benchmarks. The consistent gains on both R2R-CE and RxR-CE indicate that explicit local geometric enhancement provides robust benefits under both goal-reaching-oriented and path-fidelity-oriented evaluation settings.

\subsection{Ablation Study}
To better understand the contribution of each design choice in our LCG module, we conduct ablation studies on the R2R-CE Val-Unseen split. We use ETP-R1\cite{ye2025etp}, in both its SFT and RFT stages, as the baseline for these component analyses.

\textbf{1) Impact of Representation Formats and Enhancement Scopes.}

We first examine whether depth redundancy can be removed directly in the 2D image domain. As shown in Table~\ref{tab:ablation_repr_scope}, both Global 2D Cropping and Local 2D Cropping lead to clear performance degradation, with Global 2D Cropping causing the most severe drop. This suggests that direct image-domain truncation is not a suitable solution: although it removes distant depth responses, it also disrupts the original view-level depth representation and introduces unnatural patterns for the visual backbone. In contrast, projecting depth into 3D space enables physically meaningful truncation before feature extraction, which better preserves local geometric structure and explains the necessity of explicit 3D geometric modeling.

We further compare global and local enhancement scopes. Global 3D enhancement brings only limited gains and may even reduce SR, indicating that indiscriminately injecting geometric features into the global graph can still interfere with local decision-making. By contrast, our Local 3D setting enhances only the ghost nodes connected to the current location and degrades outdated nodes back to their original representations. This setting achieves the best overall results in both SFT and RFT stages, demonstrating that localized enhancement effectively mitigates the dilution caused by global map reasoning and helps the planner focus on the current frontier.

\begin{table}[t]
\caption{Ablation Study on Geometric Representation Formats and Enhancement Scopes}
\label{tab:ablation_repr_scope}
\centering
{\tblfont
\setlength{\tabcolsep}{2.2pt}
\renewcommand{\arraystretch}{\tblstretch}
\begin{tabular}{@{}c c c c c c c c@{}}
\hline
\multirow{2}{*}{\makecell{Training\\Phase}} &
\multirow{2}{*}{\makecell{Depth\\Crop.}} &
\multirow{2}{*}{\makecell{Point-cloud\\Enh.}} &
\multicolumn{5}{c}{Val-Unseen} \\
\cline{4-8}
 &  &  & TL & NE$\downarrow$ & OSR$\uparrow$ & SR$\uparrow$ & SPL$\uparrow$ \\
\hline
\multirow{5}{*}{\parbox[c]{0.85cm}{\centering SFT}}
 & --     & --     & 13.34 & 4.11 & 68.62 & 63.24 & 54.30 \\
 & Global & --     & 14.00 & 4.93 & 55.74 & 51.82 & 42.92 \\
 & Local  & --     & 13.62 & 4.50 & 62.53 & 58.08 & 49.27 \\
 & --     & Global & 12.80 & 4.10 & 68.95 & 63.13 & 54.69 \\
 & --     & Local  & \textbf{12.56} & \textbf{4.07} & \textbf{69.28} & \textbf{63.89} & \textbf{55.21} \\
\hline
\multirow{5}{*}{\parbox[c]{0.85cm}{\centering RFT}}
 & --     & --     & 12.74 & 3.94 & 71.45 & 65.31 & 55.77 \\
 & Global & --     & 12.79 & 4.79 & 58.24 & 53.45 & 45.02 \\
 & Local  & --     & 13.01 & 4.39 & 64.93 & 59.71 & 50.77 \\
 & --     & Global & 11.94 & 3.99 & 69.82 & 64.11 & 55.97 \\
 & --     & Local  & \textbf{11.84} & \textbf{3.93} & 70.69 & 65.14 & \textbf{56.85} \\
\hline
\end{tabular}
}
\end{table}

\begin{table}[t]
\caption{Ablation Study on Fusion Strategies}
\label{tab:ablation_fusion}
\centering
{\tblfont
\setlength{\tabcolsep}{\tblcolsep}
\renewcommand{\arraystretch}{\tblstretch}
\begin{tabular}{@{}c c c c c c c@{}}
\hline
\multirow{2}{*}{\makecell{Training\\Phase}} &
\multirow{2}{*}{\makecell{Fusion\\Strategy}} &
\multicolumn{5}{c}{Val-Unseen} \\
\cline{3-7}
 &  & TL & NE$\downarrow$ & OSR$\uparrow$ & SR$\uparrow$ & SPL$\uparrow$ \\
\hline
\multirow{3}{*}{\parbox[c]{0.85cm}{\centering SFT}}
 & --       & 13.34 & 4.11 & 68.62 & 63.24 & 54.30 \\
 & Direct   & 12.20 & 4.15 & 68.57 & 62.59 & 54.21 \\
 & Weighted & \textbf{12.56} & \textbf{4.07} & \textbf{69.28} & \textbf{63.89} & \textbf{55.21} \\
\hline
\multirow{3}{*}{\parbox[c]{0.85cm}{\centering RFT}}
 & --       & 12.74 & 3.94 & 71.45 & 65.31 & 55.77 \\
 & Direct   & 11.59 & 4.08 & 69.11 & 64.22 & 56.02 \\
 & Weighted & 11.84 & \textbf{3.93} & 70.69 & 65.14 & \textbf{56.85} \\
\hline
\end{tabular}
}
\end{table}

\textbf{2) Impact of Fusion Strategies.}

We further study how to integrate the extracted 3D geometric features with the pre-trained topological representation. As shown in Table~\ref{tab:ablation_fusion}, Direct Fusion consistently underperforms the baseline, indicating that naive feature addition can disturb the original feature space and weaken the learned cross-modal alignment. In contrast, the proposed Weighted Fusion achieves the best overall performance by using a learnable scalar to control the contribution of geometric cues. This result confirms that local 3D features should be introduced as a calibrated complement to the original ghost-node representation, rather than being directly injected without modulation.

\begin{table}[t]
\caption{Ablation Study on Truncation Depth and Downsampling Points}
\label{tab:ablation_trunc_depth_points}
\centering
{\tblfont
\setlength{\tabcolsep}{2.15pt}
\renewcommand{\arraystretch}{\tblstretch}
\begin{tabular}{@{}c c c c c c c c c@{}}
\hline
\multirow{2}{*}{Model} &
\multirow{2}{*}{Depth} &
\multirow{2}{*}{Points} &
\multicolumn{6}{c}{Val-Unseen} \\
\cline{4-9}
 &  &  & TL & NE$\downarrow$ & OSR$\uparrow$ & SR$\uparrow$ & SPL$\uparrow$ & Avg$_t$ \\
\hline
\multirow{5}{*}{\parbox[c]{1.05cm}{\centering LCG-\\ETP-R1\\(RFT)}}
 & 3.0  & 1024 & 11.84 & 3.94 & 70.47 & 64.87 & 56.69 & 1.529 \\
 & 3.0  & 512  & 11.86 & 3.94 & 70.47 & 64.82 & 56.67 & 0.889 \\
 & 3.0  & 256  & \textbf{11.84} & \textbf{3.93} & \textbf{70.69} & \textbf{65.13} & \textbf{56.85} & \textbf{0.584} \\
 & None & 256  & 12.40 & 4.37 & 67.86 & 61.50 & 52.40 & 0.573 \\
 & 5.0  & 256  & 11.83 & 3.94 & 69.66 & 64.43 & 55.92 & 0.547 \\
\hline
\end{tabular}
}
\end{table}

\textbf{3) Impact of Truncation Depth and Downsampling Points}

We further analyze how the truncation depth and sampling density affect LCGNav. As shown in Table~\ref{tab:ablation_trunc_depth_points}, the 3m truncation setting achieves the best overall performance, while no truncation or a larger 5m range introduces more distant geometry and leads to weaker results. This supports our assumption that geometric enhancement should focus on the agent's immediate reachable space rather than the full depth range. We also compare different point numbers under the 3m setting. Increasing the point count to 512 or 1024 brings only marginal performance changes, but substantially increases the inference time. Therefore, we use 3m truncation and 256 sampled points as the default configuration, which provides a practical balance between local geometric fidelity and computational efficiency.

\textbf{4) Qualitative Analysis of Solved Navigation Failures}

To provide an intuitive understanding of how the LCG module affects spatial reasoning, we conduct a qualitative analysis of topological trajectories. As illustrated in Figure 4, we select specific episodes where the baseline model (ETPNav\cite{an2024etpnav}) fails and compare them with the corresponding trajectories of the geometrically enhanced model (LCG-ETPNav). The examples highlight three common failure modes: incorrect initial direction selection, wrong branching at intersections, and repeated local looping. In all three cases, the added local geometric cues help the agent better distinguish reachable frontiers from misleading alternatives, while the state degradation strategy reduces interference from previously visited locations. As a result, the enhanced model exhibits more stable initial decisions, fewer redundant detours at intersections, and less looping behavior in visually ambiguous areas.

\begin{figure}[!t]
\centering
\includegraphics[width=3.5in]{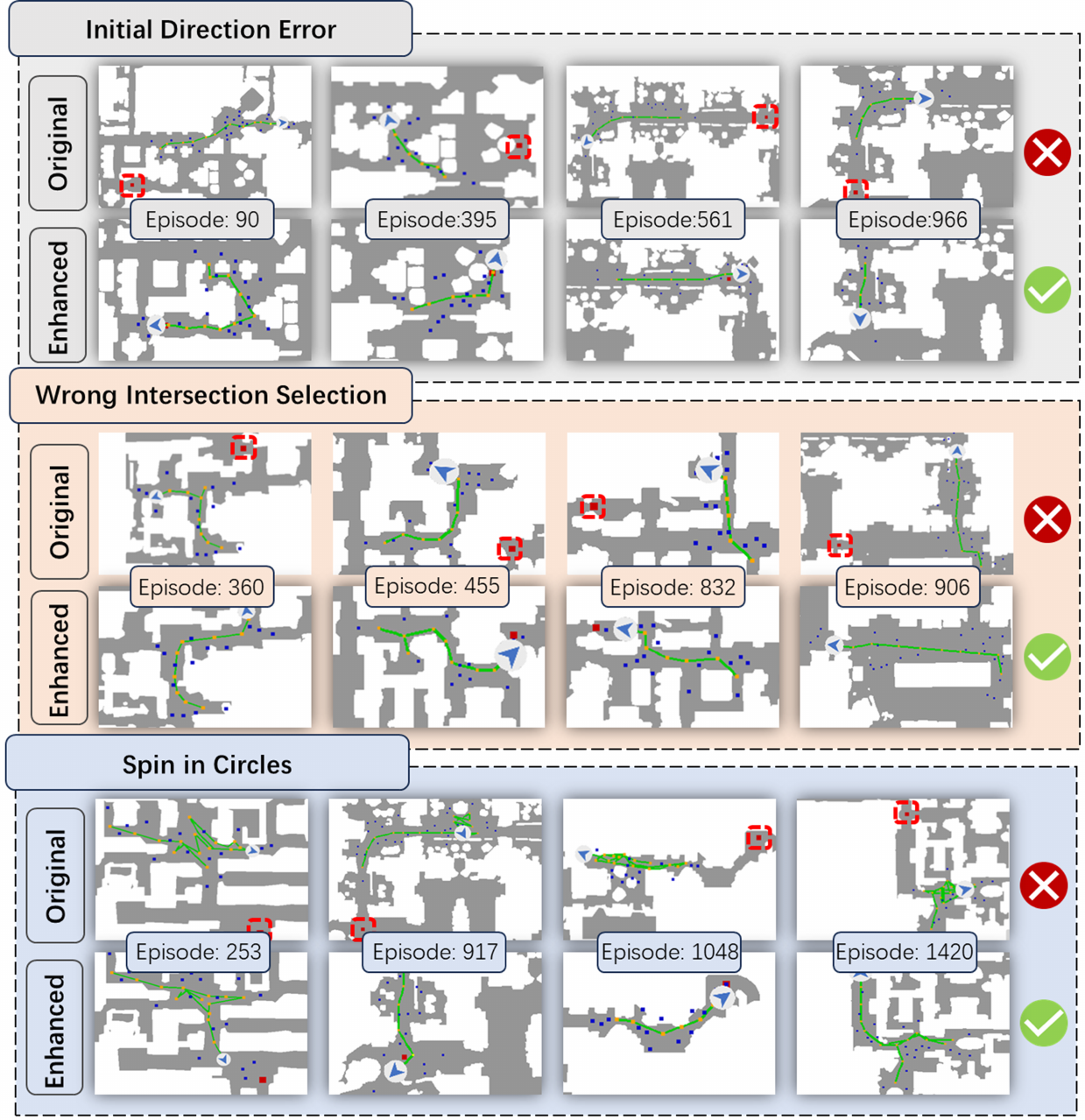}
\caption{ Qualitative comparison of topological navigation trajectories. We visualize specific episodes where the standard ETPNav baseline fails (top rows) and compare them with the corresponding trajectories after integrating the LCG module (bottom rows). The comparative analysis highlights three primary failure modes: Initial Direction Error, Wrong Intersection Selection, and Spin in Circles.}
\label{fig_4}
\end{figure}

\section{CONCLUSIONS}

In this paper, we propose LCGNav, a modular geometric enhancement framework for online topological vision-language navigation. By converting candidate depth views into truncated 3D point clouds and introducing dimension-preserving local fusion with transient state degradation, LCGNav improves local geometric perception while maintaining compatibility with existing topological planners. Experiments on both R2R-CE and RxR-CE show that LCGNav generalizes well across multiple architectures and improves multiple key metrics on representative baselines. In particular, when integrated with ETP-R1, LCGNav achieves the best performance among the compared online topological methods on the val-unseen splits of both benchmarks. These results suggest that explicit local geometric enhancement is a practical and generalizable direction for improving topological planning in VLN-CE. Future work will consider sensor noise and real-robot deployment to further evaluate the robustness and practical navigation performance of LCGNav.

\balance
\bibliographystyle{IEEEtran}
\bibliography{reference}

\end{document}